\pdfoutput=1

\documentclass[11pt]{article}

\usepackage[final]{acl}
\usepackage{times}
\usepackage{latexsym}

\usepackage[T1]{fontenc}

\usepackage[utf8]{inputenc}

\usepackage{microtype}

\usepackage{inconsolata}

\usepackage{graphicx}

\usepackage{amsmath,amsfonts,bm}

\def\eqref#1{equation~\ref{#1}}

\def\1{\bm{1}}

\DeclareMathAlphabet{\mathsfit}{\encodingdefault}{\sfdefault}{m}{sl}
\SetMathAlphabet{\mathsfit}{bold}{\encodingdefault}{\sfdefault}{bx}{n}

\newcommand{\E}{\mathbb{E}}

\usepackage{hyperref}
\usepackage{url}
\usepackage{pifont}
\usepackage{colortbl}
\usepackage{amsmath, amssymb}
\usepackage{enumitem}
\usepackage{graphicx}
\usepackage{subcaption}
\usepackage{booktabs}
\usepackage{algorithm}
\usepackage{algpseudocode}
\usepackage{tcolorbox}

\tcbset{
  myaddedbox/.style={
    colback=blue!5,
    colframe=blue!75,
    fonttitle=\bfseries,
    boxrule=0.5mm,
    arc=2mm,
    left=2mm,
    right=2mm,
    top=2mm,
    bottom=2mm
  }
}
\tcbset{
  simplebox/.style={
    colback=white,     
    colframe=blue!75,    
    boxrule=0.4pt,     
    sharp corners,     
    left=1mm,          
    right=1mm,
    top=1mm,
    bottom=1mm
  }
}
\usepackage{listings}

\lstdefinelanguage{Markdown}{
  keywords={},
  sensitive=false,
}

\title{To Code or not to Code? Adaptive Tool Integration for Math Language Models via Expectation-Maximization}

\author{
  \textbf{Haozhe Wang}
  \textsuperscript{\rm 1}, %
  \textbf{Long Li}
  \textsuperscript{\rm 2}, %
  \textbf{Chao Qu} \textsuperscript{\rm 2}, %
  \textbf{Fengming Zhu}\textsuperscript{\rm 1}, \\ 
  \textbf{Weidi Xu}\textsuperscript{\rm 2}, %
  \textbf{Wei Chu} \textsuperscript{\rm 2}, %
  \textbf{Fangzhen Lin}\textsuperscript{\rm 1}\footnotemark[2] \\
  Hong Kong University of Science and Technology\textsuperscript{\rm 1}, INFLY Tech\textsuperscript{\rm 2}
}

\definecolor{LightCyan}{rgb}{0.88,1,1}
\newcommand{\xmark}{\ding{55}} 

\newcommand{\up}{$\uparrow$}

\begin{document}

\maketitle

\begin{abstract}
Recent advances in mathematical problem-solving with language models (LMs) integrate chain-of-thought (CoT) reasoning and code execution to harness their complementary strengths. However, existing hybrid frameworks exhibit a critical limitation: they depend on externally dictated instructions or rigid code-integration templates, lacking metacognitive awareness -- the capacity to dynamically evaluate intrinsic capabilities and autonomously determine when and how to integrate tools. This rigidity motivates our study of autonomous code integration, enabling models to adapt tool-usage strategies as their reasoning abilities evolve during training.

While reinforcement learning (RL) shows promise for boosting LLM reasoning at scale (e.g., DeepSeek-R1), we demonstrate its inefficiency in learning autonomous code integration due to inadequate exploration of the vast combinatorial space of CoT-code interleaving patterns.  To address this challenge, we propose a novel
Expectation-Maximization (EM) framework that
synergizes structured exploration (E-step) with off-policy RL optimization (M-step), creating a self-reinforcing cycle between metacognitive tool-use decisions and evolving capabilities.  Experiments reveal our method achieves superior results through improved exploration. Notably, our 7B model improves over 11\% on MATH500 and 9.4\% on AIME without o1-like CoT. Code, models and data are released on \href{https://github.com/HaozheH3/AutoCode}{https://github.com/HaozheH3/AutoCode}.
\end{abstract}

\section{Introduction}\label{sec1}

Large Language Models (LLMs) have demonstrated remarkable performance across various domains~\citep{app, gpt4, llama3, gemini, yang2024qwen2}. Yet, solving complex mathematical problems still remains challenging, as the task requires hybrid skills in abstract reasoning, symbolic manipulation, and precise numerical computation~\citep{pal, mammoth, tora, htl}. 
Current approaches adopt two complementary paradigms: (1) \emph{chain-of-thought (CoT) reasoning}, which decomposes problems into intermediate reasoning steps~\citep{cot, metamath}, and (2) \emph{external tool integration}, where models generate code snippets to offload computations to interpreters or symbolic solvers~\citep{openmath, mammoth}. While CoT reasoning excels at semantic parsing and stepwise logic, its reliance on token-level autoregressive generation often propagates numerical errors. Conversely, tool-based approaches ensure computational precision but suffer from a semantic-to-symbolic translation gap, where even minor syntactic errors or contextual misinterpretations disrupt execution~\citep{htl}.

\begin{figure}[t]
    \centering 
    \includegraphics[width=\linewidth]{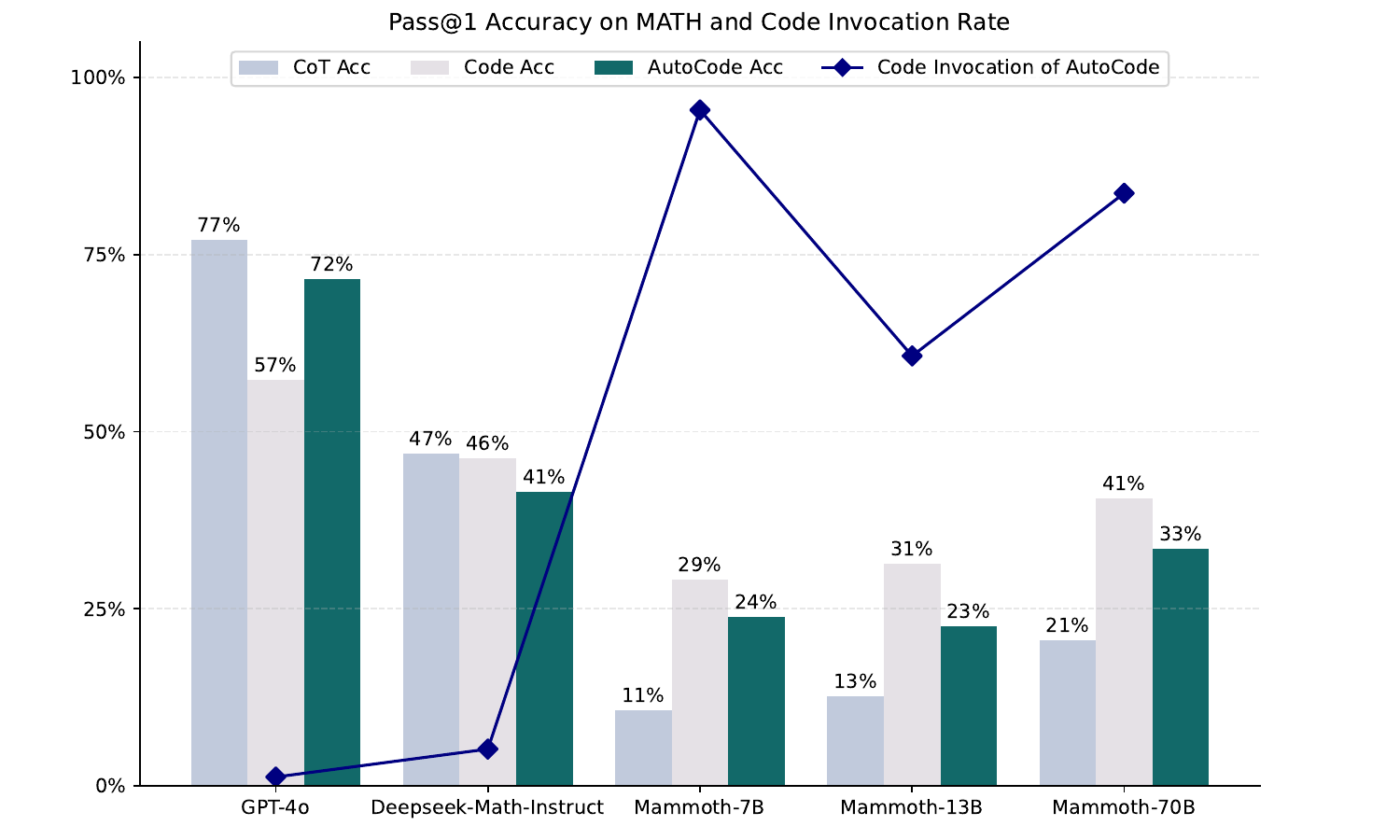} 
    \caption{\small \textbf{Analysis of existing tool-integrated math LLMs.} \small While tool integration can in principle bring complementary benefits to CoT reasoning, existing models show critical rigidity when incorporating code, lacking the metacognitive capacity to earn the synergistic benefits from code integration.  } 
    \label{fig_analysis} 
    \vspace{-0.2cm}
\end{figure}
Recent hybrid frameworks like Mammoth~\citep{mammoth}, Deepseek-Math~\citep{tora, dsmath}, and Qwen-2.5-Math~\citep{qwen25} attempt to combine these paradigms through interleaved CoT-code reasoning. However, as our empirical analysis reveals (Fig.~\ref{fig_analysis}), current methods exhibit a critical rigidity: they either default to CoT reasoning unless explicitly prompted for code generation or adhere to static templates for tool invocation.  We trace this limitation to prevailing supervised fine-tuning (SFT) paradigms that condition models to (1) passively follow user instructions (e.g., "Let’s write a Python program"~\citep{mammoth}), (2) replicate fixed code-integration patterns from curated datasets~\citep{qwen25}, or (3) imitate teacher-forced tool-use trajectories~\citep{tora, dsmath}.  Consequently, LLMs lack \emph{metacognitive awareness} -- the capacity to dynamically evaluate their intrinsic capabilities against problem contexts and autonomously determine when and how to integrate tools. This deficiency motivates our central research question:

\vspace{0.1in}
\emph{How can mathematical LLMs learn autonomous code integration (AutoCode) that optimally complements their inherent reasoning capabilities?}
\vspace{0.1in}

Reinforcement learning (RL) offers a promising pathway by optimizing policies through self-generated trajectories, as evidenced by recent successes like DeepSeek R1~\citep{dsr1}. However, we empirically observe that standard RL methods is inefficient in learning autonomous code integration (\textit{AutoCode}) strategies (see Sec.~\ref{sec_ablation}). This stems from RL's tendency to exploit local policy neighborhoods, thereby insufficiently exploring the vast combinatorial space of potential CoT-code interleaving patterns. Such myopic exploration constrains the discovery of high-reward reasoning paths that judiciously blend both modalities, particularly as the model’s reasoning capabilities evolve during training.

To address this challenge, we propose a novel Expectation-Maximization (EM) framework that synergizes guided exploration with policy optimization. Our key innovation lies in formulating code-integration decisions as latent variables within an EM paradigm, creating a self-reinforcing cycle: the E-step identifies high-potential code-integration decisions through guided exploration, while the M-step optimizes policy parameters for joint metacognitive tool-usage and reasoning. 

This dual mechanism enables models to adapt tool-use strategies as their capabilities evolve during training. Practically, we achieve efficiency through two design choices: (1) an offline data curation step (E-step) that prioritizes high-return code invocation decisions through guided exploration, and (2) an off-policy RL optimization step (M-step) that jointly improves tool-usage and reasoning. This approach offers enhanced control and efficiency compared to standard RL, which is particularly beneficial for resource-constrained companies or researchers.

Extensive experiments demonstrate that our method (a) preserves higher training efficiency while achieving better performance, and (b) learns intelligent code integration strategies that achieves higher accuracy than either CoT or code prompted in isolation. Notably, our show consistent improvements across different benchmarks, raising MATH500 from 60.4\% to 71.4\%. 

Our contribution is summarized as follows:
(1) We diagnose a critical gap in mathematical LLM -- the inability to autonomously integrate tools based on metacognitive awareness -- and demonstrate standard RL’s inefficiency in addressing it.
(2) We propose a novel EM-based framework that jointly adapts the tool-usage strategies with evolving reasoning abilities, with a simple yet efficient implementation. 
(3) We demonstrate superior results in both training efficiency and accuracy on challenging benchmarks.

\section{Background}


\begin{figure}[t]
    \centering 
    \includegraphics[width=\linewidth]{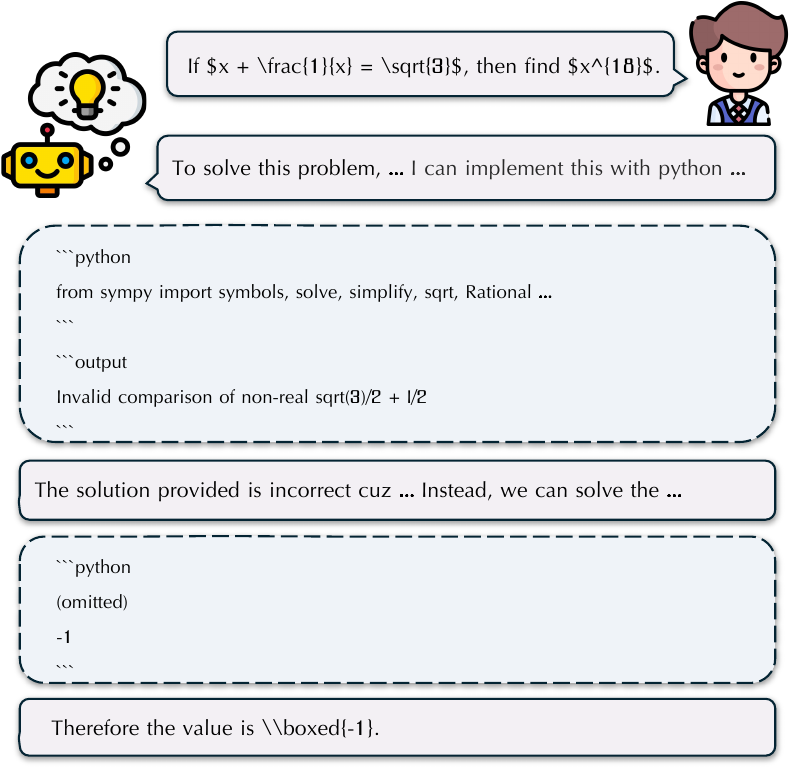} 
    \caption{\small \textbf{Example of Autonomous Code Integration.} \small We aim to enable LLMs to determine tool-usage strategies
based on their own capability boundaries. In the example, the model write code to solve the problem that demand special tricks, strategically bypassing its inherent limitations.} 
    \label{fig_example}
    \vspace{-0.2cm}
\end{figure}
\textbf{Problem Statement.} Modern tool-augmented language models address mathematical problems \( x_q \in \mathcal{X}_Q \) by generating step-by-step solutions that interleave natural language reasoning with executable Python code (Fig.~\ref{fig_example}). Formally, given a problem \( x_q \), a model \( \mathcal{M}_\theta \) iteratively constructs a solution \( y_a = \{y_1, \dots, y_T\} \) by sampling components \( y_t \sim p(y_t | y_{<t}, x_q) \), where \( y_{<t} \) encompasses both prior reasoning steps, code snippets and execution results \( \mathbf{e}_t \) from a Python interpreter. The process terminates upon generating an end token, and the solution is evaluated via a binary reward \( r(y_a,x_q) = \mathbb{I}(y_a \equiv y^*) \) indicating equivalence to the ground truth \( y^* \). The learning objective is formulated as:
\[
\max_{\theta} \mathbb{E}_{x_q \sim \mathcal{X}_Q} \left[r(y_a, x_q) \right]
\]

\noindent\textbf{Challenge and Motivation.} Developing autonomous code integration (AutoCode) strategies poses unique challenges, as optimal tool-usage behaviors must dynamically adapt to a model's intrinsic capabilities and problem-solving contexts. While traditional supervised fine-tuning (SFT) relies on imitation learning from expert demonstrations, this paradigm fundamentally limits the emergence of self-directed tool-usage strategies. Unfortunately, current math LLMs predominantly employ SFT to orchestrate tool integration~\citep{mammoth, tora, dsmath, htl}, their rigid adherence to predefined reasoning templates therefore struggles with the dynamic interplay between a model’s evolving problem-solving competencies and the adaptive tool-usage strategies required for diverse mathematical contexts.

Reinforcement learning (RL) offers a promising alternative by enabling trial-and-error discovery of autonomous behaviors. Recent work like DeepSeek-R1~\citep{dsr1} demonstrates RL's potential to enhance reasoning without expert demonstrations. However, we observe that standard RL methods (e.g., PPO~\cite{ppo}) suffer from a critical inefficiency (see Sec.~\ref{sec_ablation}): Their tendency to exploit local policy neighborhoods leads to insufficient exploration of the vast combinatorial space of code-integrated reasoning paths, especially when only given a terminal reward in mathematical problem-solving.

To bridge this gap, we draw inspiration from human metacognition -- the iterative process where learners refine tool-use strategies through deliberate exploration, outcome analysis, and belief updates. A novice might initially attempt manual root-finding via algebraic methods, observe computational bottlenecks or inaccuracies, and therefore prompting the usage of calculators. Through systematic reflection on these experiences, they internalize the contextual efficacy of external tools, gradually forming stable heuristics that balance reasoning with judicious tool invocation.

To this end, \emph{our focus diverges from standard agentic tool-use frameworks~\citep{agentr}}, which merely prioritize successful tool execution. Instead, \emph{we aim to instill \emph{human-like metacognition} in LLMs, enabling them to (1) determine tool-usage based on their own capability boundaries (see the analysis in Sec.~\ref{sec_ablation}), and (2) dynamically adapt tool-usage strategies as their reasoning abilities evolve (via our EM framework).}


\section{Methodology}

\begin{figure*}[t]
    \centering
    \includegraphics[width=\linewidth]{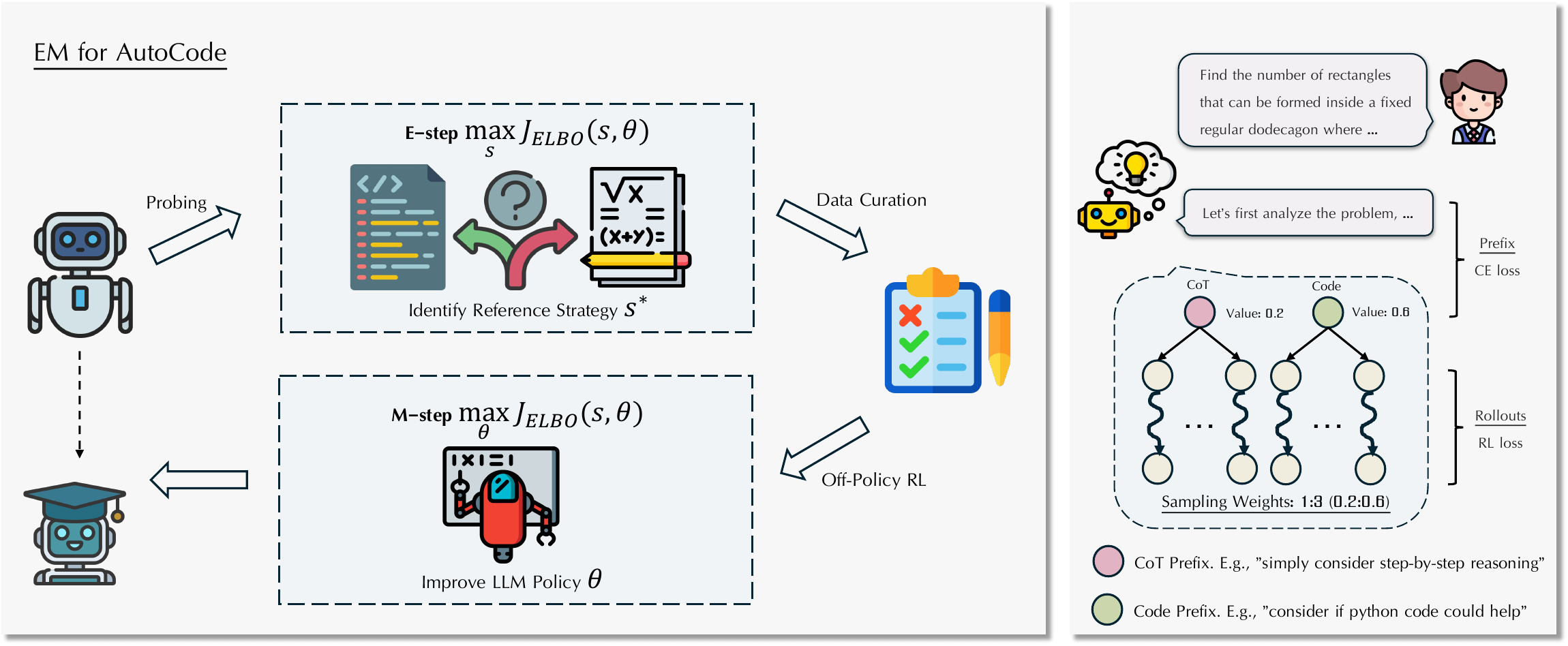}
    \caption{\small \textbf{Method Overview.} \small (Left) shows an overview for the EM framework, which alternates between finding a reference strategy for guided exploration (E-step) and off-policy RL (M-step). (Right) shows the data curation for guided exploration. We generate \(K\) rollouts, estimate values of code-triggering decisions and subsample the initial data with sampling weights per Eq.~\ref{eq_sampling}.}
    \label{fig_overview}
\end{figure*}

Inspired by human metacognitive processes, we introduce an Expectation-Maximization (EM) framework that trains LLMs for autonomous code integration (AutoCode) through alternations (Fig.~\ref{fig_overview}):

\begin{enumerate}[leftmargin=0.5cm,topsep=1pt,itemsep=0pt,parsep=0pt]
    \item \emph{Guided Exploration (E-step):} Identifies high-potential code-integrated solutions by systematically probing the model's inherent capabilities.
\item \emph{Self-Refinement (M-step):} Optimizes the model's tool-usage strategy and chain-of-thought reasoning using curated trajectories from the E-step.
\end{enumerate}

\subsection{The EM Framework for AutoCode}

A central challenge in AutoCode lies in the code triggering decisions, represented by the binary decision \(c \in \{0, 1\}\).  While supervised fine-tuning (SFT) suffers from missing ground truth for these decisions, standard reinforcement learning (RL) struggles with the combinatorial explosion of code-integrated reasoning paths. Our innovation bridges these approaches through systematic exploration of both code-enabled (\(c=1\)) and non-code (\(c=0\)) solution paths, constructing reference decisions for policy optimization.

We formalize this idea within a maximum likelihood estimation (MLE) framework. Let \( P (r=1 | x_q;\theta\) denote the probability of generating a correct response to query \( x_q \) under model \(\mathcal{M}_\theta\). Our objective becomes:
\begin{align}
    \mathcal{J}_{\mathrm{MLE}}(\theta) \doteq \log P(r=1 | x_q; \theta) \label{eq_mle}
\end{align}
This likelihood depends on two latent factors: (1) the code triggering decision \(\pi_\theta(c | x_q)\) and (2) the solution generation process \(\pi_\theta(y_a | x_q, c)\). Here, for notation-wise clarity, we consider  code-triggering decision at a solution's beginning (\( c\) following \(x_q\) immediately). We show generalization to mid-reasoning code integration in Sec.~\ref{sec_impl}.

The EM framework provides a principled way to optimize this MLE objective in the presence of latent variables~\cite{prml}. We derive the evidence lower bound (ELBO): \( \mathcal{J}_{\mathrm{ELBO}}(s, \theta) \doteq \)
\begin{align}
    \mathbb{E}_{s(c | x_q)}\left[\log \frac{\pi_\theta(c | x_q) \cdot P(r=1 | c, x_q; \theta)}{s(c | x_q)}\right] 
     \label{eq_elbo}
\end{align}
where \(s(c | x_q)\) serves as a surrogate distribution approximating optimal code triggering strategies. It is also considered as the reference decisions for code integration. 

\noindent\textbf{E-step: Guided Exploration}  computes the reference strategy \(s(c | x_q)\) by maximizing the ELBO, equivalent to minimizing the KL-divergence: \( \max_s \mathcal{J}_{\mathrm{ELBO}}(s, \theta) = \)
\begin{align}
     - \mathrm{D_{KL}}\left(s(c | x_q) \| P(r=1, c | x_q; \theta)\right) \label{eq_estep}
\end{align}

The reference strategy \(s(c | x_q)\) thus approximates the posterior distribution over code-triggering decisions \(c\) that maximize correctness, i.e., \(P(r=1, c | x_q; \theta)\).  Intuitively, it guides exploration by prioritizing decisions with high potential: if decision \(c\) is more likely to lead to correct solutions, the reference strategy assigns higher probability mass to it, providing guidance for the subsequent RL procedure.

\noindent\textbf{M-step: Self-Refinement } updates the model parameters \(\theta\) through a composite objective:
\begin{multline}
\max_\theta \mathcal{J}_{\mathrm{ELBO}}(s, \theta) =\mathbb{E}_{\substack{c \sim s(c|x_q) \\ y_a \sim \pi_\theta(y_a|x_q, c)}} \Big[ r(x_q, y_a) \Big] \\- \mathcal{CE}\Big(s(c|x_q) \,\|\, \pi_\theta(c|x_q)\Big)\label{eq_mstep}
\end{multline}
The first term implements reward-maximizing policy gradient updates for solution generation, while while the second aligns native code triggering with reference strategies through cross-entropy minimization (see Fig.~\ref{fig_overview} for an illustration of the optimization). This dual optimization jointly enhances both tool-usage policies and reasoning capabilities.

\subsection{Practical Implementation}\label{sec_impl}
In the above EM framework, we alternate between finding a reference strategy \( s \) for code-triggering decisions  in the E-step, and perform reinforcement learning under the guidance from \( s \) in the M-step. We implement this framework through an iterative process of offline data curation and off-policy RL.

\noindent\textbf{Offline Data Curation.} We implement the E-step through Monte Carlo rollouts and subsampling. For each problem \(x_q\), we estimate the reference strategy as an energy distribution: 
\begin{equation}
    s^\ast(c | x_q)  = \frac{\exp\left(\alpha\cdot \pi_\theta(c | x_q) Q(x_q,c;\theta)\right)}{Z(x_q)}.\label{eq_sampling}
\end{equation}
where \( Q(x_q,c;\theta)\) estimates the expected value through \( K \) rollouts per decision, \(\pi_\theta(c|x_q) \) represents the model's current prior and the \( Z(x_q) \) is the partition function to ensure normalization. Intuitively, the strategy will assign higher probability mass to the decision \( c \) that has higher expected value \( Q(x_q,c;\theta)\) meanwhile balancing its intrinsic preference \( \pi_\theta(c|x_q)\). 

Our curation pipeline proceeds through: 
\begin{itemize}[leftmargin=0.5cm,topsep=1pt,itemsep=0pt,parsep=0pt]
\item Generate \(K\) rollouts for \(c=0\) (pure reasoning) and \(c=1\) (code integration), creating candidate dataset \(\mathcal{D}\).  
\item Compute \(Q(x_q,c)\) as the expected success rate across rollouts for each pair \((x_q,c)\).  
\item Subsample \(\mathcal{D}_{\text{train}}\) from \(\mathcal{D}\) using importance weights according to Eq.~\ref{eq_sampling}.  
\end{itemize}

To explicitly probe code-integrated solutions, we employ prefix-guided generation -- e.g., prepending prompts like \texttt{``Let’s first analyze the problem, then consider if python code could help''} -- to bias generations toward free-form code-reasoning patterns.

 This pipeline enables guided exploration by focusing on high-potential code-integrated trajectories identified by the reference strategy, contrasting with standard RL’s reliance on local policy neighborhoods. As demonstrated in Sec.~\ref{sec_ablation}, this strategic data curation significantly improves training efficiency by shaping the exploration space.

\noindent\textbf{Off-Policy RL.}
To mitigate distributional shifts caused by mismatches between offline data and the policy, we optimize a clipped off-policy RL objective. The refined M-step (Eq.~\ref{eq_mstep}) becomes:
\begin{multline}
    \underset{(x_q,y_a)}{\mathbb{E}}\left[
\text{clip}\left(\frac{\pi_\theta(y_a|x_q)}{\pi_{\text{ref}}(y_a|x_q)},1-\epsilon,1+\epsilon\right)\cdot A\right]
\\-\mathbb{E}_{(x_q,c)}\Big[\log \pi_\theta(c|x_q) \Big]\label{eq_finalm}
\end{multline}
where  \( (x_q, c, y_a) \) is sampled from the dataset \( \mathcal{D}_{\text{train}} \). The importance weight \(\frac{\pi_\theta(y_a|x_q)}{\pi_{\text{ref}}(y_a|x_q)}\) accounts for off-policy correction with PPO-like clipping. The advantage function \(A(x_q,y_a)\) is computed via query-wise reward normalization~\cite{ppo}. 

\noindent\textbf{Generalizing to Mid-Reasoning Code Integration.} Our method extends to mid-reasoning code integration by initiating Monte Carlo rollouts from partial solutions \((x_q, y_{<t})\). Notably, we observe emergence of mid-reasoning code triggers after initial warm-up with prefix-probed solutions. Thus, our implementation requires only two initial probing strategies: explicit prefix prompting for code integration and vanilla generation for pure reasoning, which jointly seed diverse mid-reasoning code usage in later iterations.

\section{Experiments}\label{sec_exp}

\begin{table*}[ht]
\caption{\small \textbf{Public SFT Data Used in this Work.} We collect public query set for AutoCode Training. After Deduplication, the total amount of query used is \(119\)K. If the base model is not trained to write code for math problems, we use the SFT annotations associated with the above queries. }
\centering
\tiny
\resizebox{\linewidth}{!}{%
\begin{tabular}{ccccc}
\toprule
\textbf{Dataset}                                     & \textbf{Size}  & \textbf{Unique Queries}     & \textbf{CoT Solutions}     & \textbf{Code Solutions}          \\
\midrule
Openmath~\cite{openmath}                      & 129917	&70002 & 25116	 & 104801                \\
Math-Instruct~\citep{mammoth}              &  237781   & 219607  & 188644     &  49137             \\
Metamath~\cite{mammoth}                  & 285000    & 161337   &285000       &    0                 \\ 
MMOS~\citep{mmos}                      & 134610    &69007    &   0     & 134610 \\
\hline
\end{tabular}
}

\label{tab_sft_data}
\end{table*}

\vspace{-0.2cm}
Our experiments investigate three key research questions:

\noindent\emph{Q1: Method Effectiveness.} How does our approach enhance performance across both in-domain and out-of-domain mathematical benchmarks compared to existing math LLMs?

\noindent\emph{Q2: Baseline Comparisons.} How does our method compare to standard RL and SFT baselines in terms of training efficiency and exploration patterns?

\noindent\emph{Q3: AutoCode Analysis.} What strategies does the model learn for code integration, and how do these strategies contribute to performance gains?

\noindent\textbf{Datasets and Benchmarks.} Our method only requires a query set for training. We collect public available queries from MATH~\citep{math} and Numina~\cite{numina}, and sample \(7K\) queries based on difficulties. We upload the collected data to the annonymous repo. For evaluation, we employ: GSM8k~\citep{gsm8k}, MATH500~\citep{math}, GaokaoMath2023~\citep{mario}, OlympiadBench~\citep{olympiad}, the American Invitational Mathematics Examination (AIME24), and the American
Mathematics Competitions (AMC23). This benchmark suite spans elementary to Olympiad-level mathematics. We adopt Pass@1 accuracy~\citep{pass1, dsr1} as our primary metric, using evaluation scripts from DeepseekMath~\citep{dsmath} and Qwen2Math~\citep{yang2024qwen2}. For competition-level benchmarks (AIME/AMC), we use 64 samples with temperature 0.6 following Deepseek R1 protocols.

\begin{table*}[tb]
\caption{\small \textbf{Effectiveness of AutoCode4Math.} The column "Code?" indicates whether code integration is involved, with \ding{72} representing autonomous determination of code integration by the model.  The improvement over code-driven inference is highlighted in the colored row, denoted as \(\Delta\).}.\label{fig_main_results}

\resizebox{\textwidth}{!}{
\begin{tabular}{lccccccc}
\toprule

\textbf{Model}  &\textbf{Code?} & \multicolumn{2}{c}{\textbf{In-domain}} & \multicolumn{4}{c}{\textbf{Out-of-domain}}\\
\cmidrule(lr){3-4} \cmidrule(lr){5-8} 
&        & GSM8K                           & MATH500                           & GaoKao         & Olympiad              & AIME24 & AMC23         \\
\midrule
\multicolumn{8}{c}{Proprietary Model} \\
\midrule
OpenAI-o1-preview~\cite{o1}     & \xmark         &                             & 85.5                          & 62.1           & 52.1                    & 44.6         & 81.8  \\
GPT-4o~\citep{gpt4}                & \xmark         & 92.9                            & 76.4                          & 67.5           & 43.3                     & 9.3      &   45.8   \\

Claude-3.5-Sonnet-1022~\citep{claude}                   & \xmark          & 95                              & 78.3                           &                &                              &  16.0          &     \\
\midrule
\multicolumn{8}{c}{Open-Source Models} \\
\midrule
Mammoth-70B~\citep{mammoth}                         & \checkmark      & 76.9                            & 41.8                           & 25.2           &                            &           \\
ToRA-70B~\citep{tora}                            & \checkmark      & 84.3                            & 49.7                           & 31.7           &                             &           \\
NuminaMath-72B~\citep{numina}                       & \checkmark      &  91.4                           & 59.2                           &  49.4              &   36.7                          &   6.5         & 40.6    \\
Mathstral-7B~\citep{mathstral}                       & \xmark          & 84.9                            & 56.6                           & 46             & 21.5                     &                \\
Mammoth-Mistral-7B~\citep{mammoth}                  & \checkmark      & 74.22  &  37.8  &   22.08     &9.63         &6.67        &  20.0              \\
NuminaMath-7B-CoT~\citep{numina}                    & \xmark          & 81.27                           &53.0    & 48.83      &22.22         &3.33     & 25.0
\\

Dart-Math-DeepSeek-7B~\cite{tong2024dartmath}  & \xmark &87.64  &50.0     & 45.45   & 18.52          &3.33       &35.0 \\
Dart-Math-Llama3-8B~\cite{tong2024dartmath}  & \xmark &82.71  &45.0        &  34.80       &   23              &    0.0      &       17.5  \\

\midrule
\multicolumn{8}{c}{AutoCode Training} \\
\midrule

Qwen2Math-Base-7B~\citep{yang2024qwen2}                       & \xmark          & 80.74                           & 48.80                          & 43.37          & 21.62                 &    6.5  & 19.8    \\

\textbf{AutoCode4Math-Qwen2}    & \ding{72}        & 88.1                   & 61.86            & 50.13 & 26.37 & 13.2 & 30.0 \\
\rowcolor{LightCyan}
\textbf{$\Delta$}                               & & 7.36\up & 13.06\up  & 6.76\up           & 4.75\up           & 6.7\up          & 10.2\up           \\
\midrule 
DeepseekMath-Instruct-7B~\citep{dsmath} & \checkmark      & 84.46              & 51.00                   & 44.68          & 20.44                  & 1.6 & 17.4          \\

\textbf{AutoCode4Math-DeepSeek} & \ding{72}        & 89.26           & 63.32      &  50.53        &  26.95       & 9.5 & 28.8 \\
\rowcolor{LightCyan}
\textbf{$\Delta$}                 & &                4.8\up  & 12.32\up  & 5.85\up           & 6.51\up           &7.9\up          & 11.4\up           \\

\midrule 
Qwen-2.5-Base-7B~\citep{qwen25} & \xmark      & 84.88              & 60.4                   & 45.45          &  30.37                & 13.2 & 39.38          \\

\textbf{AutoCode4Math-Qwen2.5} & \ding{72}        &   89.12       & 71.4            &   51.69 & 32.6       & 22.6 & 45.18 \\
\rowcolor{LightCyan}
\textbf{$\Delta$}                 &                 & 4.24\up  & 11.0\up  & 6.24\up           & 2.23\up           & 9.4\up          & 5.8\up           \\
\bottomrule
\end{tabular}
}
\end{table*}

\noindent\textbf{Baselines and Implementation.} 
We compare against three model categories: \begin{itemize}[leftmargin=0.5cm,itemsep=0pt,parsep=0pt]
\item Proprietary models: o1~\cite{o1}, GPT-4~\citep{gpt4} and Claude~\citep{claude}
\item Recent math-specialized LMs: NuminaMath~\citep{numina}, Mathstral~\citep{mathstral}, Mammoth~\citep{mammoth}, ToRA~\citep{tora}, DartMath~\cite{tong2024dartmath}. We do not compare with models that rely on test-time scaling, such as MCTS or long CoT. 
\item Foundation models enhanced with our method: Qwen2Math~\citep{yang2024qwen2}, DeepseekMath~\citep{dsmath} and Qwen-2.5~\cite{qwen25}.
\end{itemize}

Our implementation uses \( K = 8 \) rollouts per query (temperature=1.0, top-p=0.9). Training completes in about 10 hours on \(8\times\) A100 (80GB) GPUs across three epochs of 7K queries. We list the collected public SFT data in Tab.~\ref{tab_sft_data}. 

\subsection{Main Results}\label{sec_main}
Notably, we observe a minimum performance gain of 11\% on the MATH500 benchmark, escalating to an impressive 9.4\% absolute improvement on the highly challenging AIME benchmark.  Across in-domain benchmarks, our method yields an average improvement of 8.9\%, and for out-of-domain benchmarks, we achieve a substantial average gain of 6.98\%. These results  validate the effectiveness of our approach across model families and problem difficulty levels.  

\subsection{Ablation Study}\label{sec_ablation}
We conduct three primary analyses: (a) comparison with standard RL and SFT baselines to validate our method's effectiveness in facilitating exploration, (b) visualization of exploration patterns to reveal limitations in the standard RL paradigam, and (c) behavioral analysis of code integration strategies. These analyses collectively demonstrate our method's benefits in facilitating guided exploration and explains how it improves performance.

\begin{figure*}[t]
    \centering
    \includegraphics[width=0.95\linewidth]{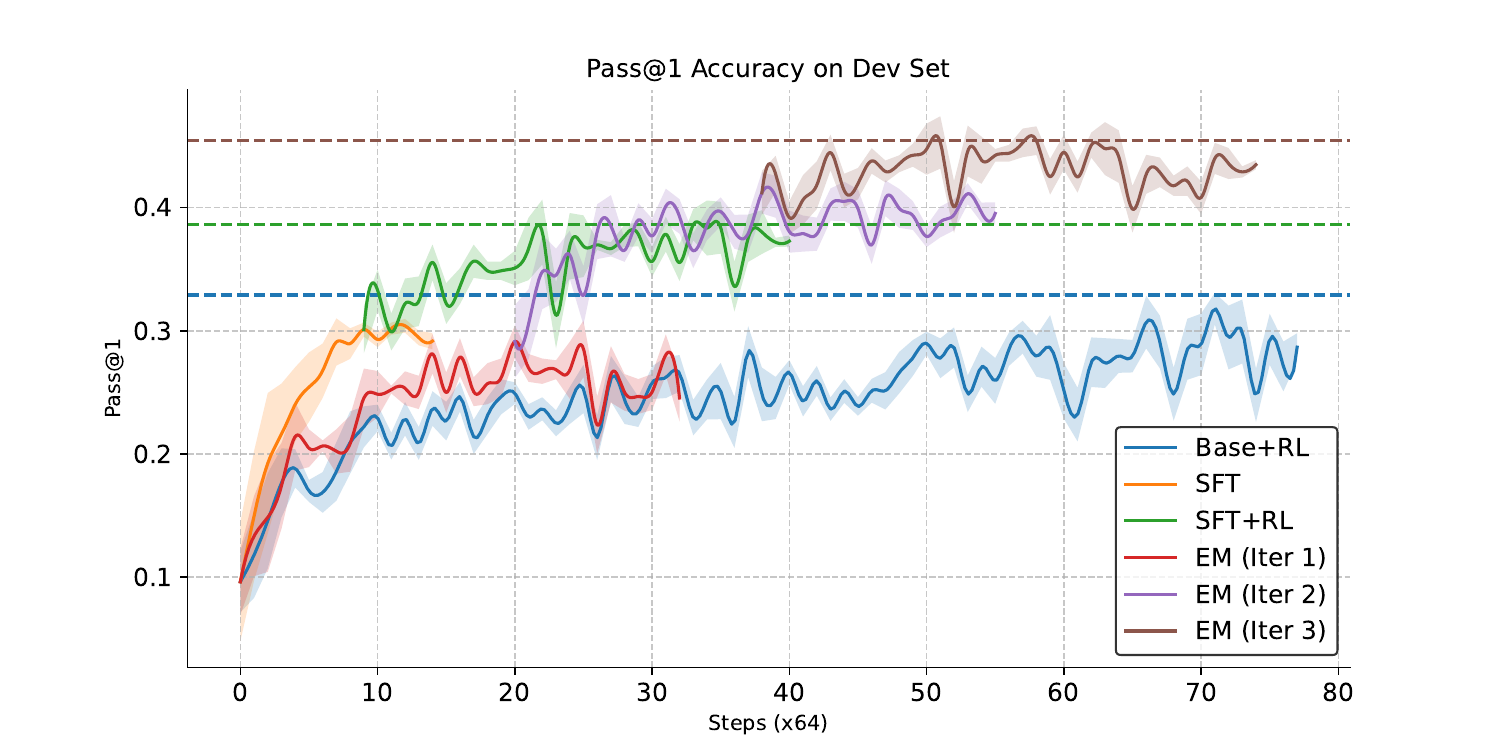}
    \caption{ \small \textbf{Training Efficiency and Convergence.} We benchmark the learning dynamics of our approach against three two training paradigms: supervised fine-tuning and reinforcement learning (RL). The Pass@1 accuracy is evaluated on an held-out dev-set. We use Qwen-2.5-Base as the base model. SFT is conducted using collected public data~\cite{openmath, mammoth}. The dashed lines indicate asymptotic performance. }\label{fig_training_efficiency}
\end{figure*}
\noindent\textbf{Training Efficiency.} We evaluated the learning dynamics of our approach in direct comparison to three established training paradigms:
\begin{itemize}[leftmargin=0.5cm,itemsep=0pt,parsep=0pt]
\item \emph{Base+RL}:  On-policy Reinforcement Learning (RL) initialized from a base model without Supervised Fine-Tuning (SFT). This follows the methodology of DeepSeek R1, designed to isolate and assess the pure effects of RL training.
\item \emph{SFT}: Supervised Fine-Tuning, the prevailing training paradigm widely adopted in current tool-integrated math Language Models (LMs).
\item \emph{SFT+RL}: Standard RL applied after SFT, serving as a conventional baseline for evaluating our EM-based RL method.
\end{itemize}

From the figure, we make the following key observations:

\begin{itemize}[leftmargin=0.5cm,itemsep=0pt,parsep=0pt]
   \item  While Reinforcement Learning directly from the base model (\emph{Base+RL}) exhibits consistent performance improvement, its training efficiency is lower than training paradigms incorporating SFT.  In addition, the model rarely explores code-integrated solutions, with the code invocation rate below 5\%. This strongly suggest that \emph{reinforcement learning tool-usage behavior from scratch is inherently inefficient}.
    \item SFT effectively provides a strong initialization point, but \emph{SFT alone exhibits limited asymptotic performance}. This suggests that SFT lacks the capacity to adapt and optimize beyond the scope of the expert demonstrations, thereby limiting further improvement. 
    \item Standard RL applied after SFT shows initial further improvement but subsequently plateaus, \emph{even after an extended training stage}.  This suggests \emph{the exploration-exploitation dilemma when applying RL for LLM post-training}: standard RL with vanilla rollout exploration tends to exploit local optima and insufficiently explores the combinatorial code-integrated trajectories.
\end{itemize}

\begin{figure}[t]
    \centering 
    \resizebox{1.\linewidth}{!}{\includegraphics[width=\linewidth]{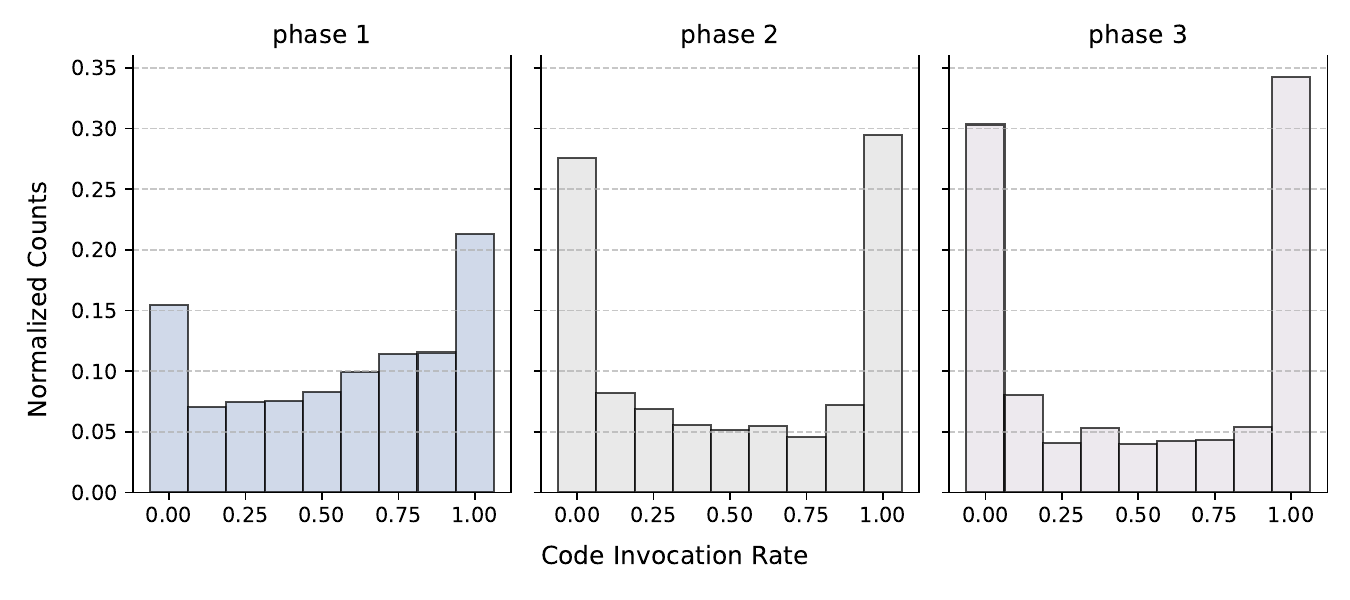} }
    \vspace{-0.2cm}
    \caption{\small \textbf{Visualization of Exploration in the SFT+RL paradigm.} \small The distribution of code invocation rates \emph{across queries} to visualize policy's exploration of code-integrated trajectories. Without external guidance, LLM tends to exploit its local policy neighborhood, concentrating code usage toward extremes as training phase evolves. } \vspace{-0.2cm}
    \label{fig_visualize_explore} 
\end{figure}

To further substantiate the exploration limitations inherent in the conventional \emph{SFT+RL} paradigm, we present a visualization of the exploration patterns. We partitioned the model-generated responses during self-exploration into three distinct training phases and analyzed the statistical distribution of code invocation rates across queries as the model's policy evolved throughout training. As depicted in Figure~\ref{fig_visualize_explore}, the distribution of code invocation progressively concentrates towards the extremes – either minimal or maximal code use – indicating the model's growing tendency to exploit its local policy neighborhood. This exploitation manifests as a focus on refining established code-triggering decisions, rather than engaging in broader exploration of alternative approaches.

These empirical observations lend strong support to our assertion that standard RL methods are susceptible to premature exploitation of the local policy space when learning AutoCode strategies. In sharp contrast, our proposed EM method facilitates a more guided exploration by sub-sampling trajectories according to the reference strategy (Sec.~\ref{sec_impl}). This enables continuous performance (evidenced in Sec.~\ref{sec_main}) and mitigating the risk of converging to suboptimal local optima (Fig.~\ref{fig_training_efficiency}).

\begin{figure}[t]
    \centering 
    \resizebox{1.\linewidth}{!}{\includegraphics[width=\linewidth]{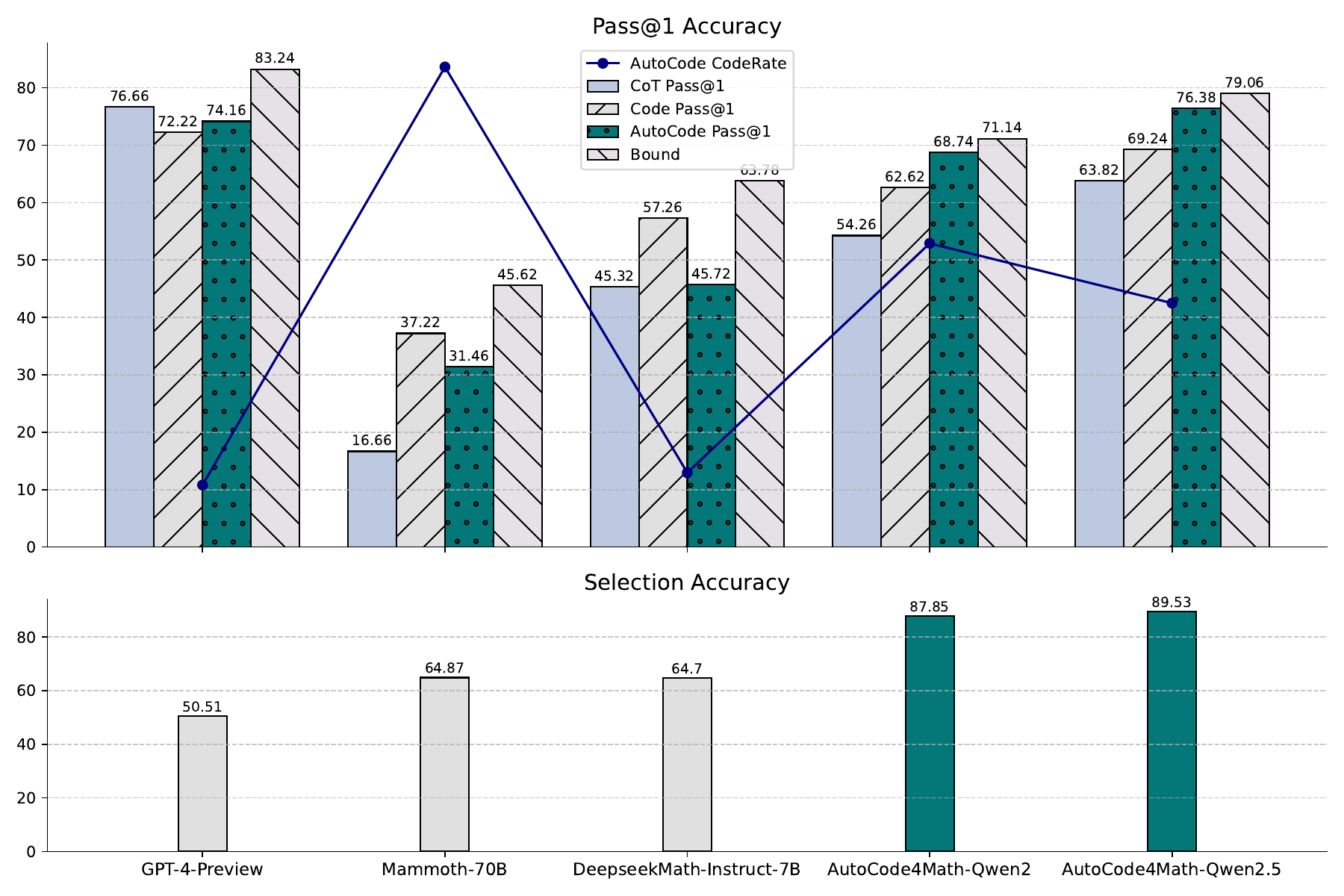}} 
    \caption{\small \textbf{Analysis of AutoCode Strategies. }\small We compare AutoCode performance against scenarios where models explicitly prompted to utilize code or CoT, and consider the union of solved queries as the bound for AutoCode performance. Existing models show inferior AutoCode performance than explicit instructed, with their AutoCode strategies close to random (50\%). Our approach consistently improves AutoCode performance, with AutoCode selection accuracy near 90\%.  } 
    \label{fig_learned_behavior} 
\end{figure}
\noindent\textbf{Analysis on Code Integration Behaviors.}
We investigated the properties of the learned code integration strategies to gain deeper insights into the mechanisms behind our method's performance gains. Our central hypothesis posits that optimal code integration unlocks synergistic performance benefits by effectively combining the strengths of CoT and code executions.  This synergy presents a "free lunch" scenario: a well-learned metacognitive tool-usage strategy can elevate overall performance, provided the model demonstrates competence in solving \emph{distinct} subsets of queries using either CoT or code execution.

To empirically validate this "free lunch" principle and demonstrate the superiority of our approach in realizing it, we benchmarked our model against baselines that inherently support both code execution and Chain-of-Thought (CoT) reasoning: GPT-4, Mammoth-70B, and DeepseekMath-Instruct-7B. Our analysis evaluated the model's autonomous decision to invoke code when not explicitly instructed on which strategy to employ. We compared this "AutoCode" performance against scenarios where models were explicitly prompted to utilize either code or CoT reasoning. We also considered the theoretical "free lunch" upper bound – the accuracy achieved by combining the successful predictions from either strategy (i.e., taking the union of queries solved by CoT or code).

As visually presented in Figure~\ref{fig_learned_behavior}, existing baseline models exhibit inferior performance in AutoCode mode compared to scenarios where code invocation is explicitly prompted, e.g., DeepseekMath-Instruct-7B shows a degradation of 11.54\% in AutoCode mode. This suggests that their AutoCode strategies are often suboptimal, performing closer to random selection between CoT and code (selection accuracy near 50\%), resulting in AutoCode falling between the performance of explicitly triggered CoT and code. In contrast, our models learn more effective code integration strategies.  AutoCode4Math-Qwen2.5, for example, improves upon explicitly code-triggered performance by 7\%, indicating a true synergistic integration of reasoning and code execution.

To quantify the effectiveness of these learned "AutoCode" strategies, we calculated the CoT/code selection accuracy. We used the outcome of explicit instruction (i.e., performance when explicitly prompted for CoT or code) as a proxy for the ground-truth optimal method selection.  Our model achieves a selection accuracy of 89.53\%, showcasing the high efficacy of the learned code integration strategy.

\section{Related Work and Discussion}
\textbf{Tool-Integrated Math LLMs.} 
Math language models adopted two major paradigms: Chain-of-Thought (CoT) reasoning and the use of external tools, such as Python programs~\citep{metamath, mammoth, openmath}. Each paradigm offers unique benefits, and recent hybrid frameworks~\citep{mammoth, tora, htl, dsmath, qwen25} increasingly seek to combine them for synergy. However, current models exhibit critical rigidity, motivating our work to realize the true metacognitive capacity that enjoys synergistic benefits of CoT and code. 

\noindent\textbf{EM for RL.} Expectation-Maximization (EM) has proven effective for maximum likelihood problems involving hidden variables, such as Expert Iteration~\citep{expertiter}, Iterative Maximum Likelihood~\citep{iml, iml1}, Meta-Reinforcement Learning~\citep{varibad, vem}, and Adversarial Games~\citep{acb}. In the context of math LLMs, the most relevant works are \citep{restem} and \citep{iml2}, which apply EM-style iterative self-training to math problem-solving. Unlike these approaches, we leverage the EM framework for guided exploration during reinforcement learning of language models.

\section{Conclusion}
Existing tool-integrated math language models lack the metacognitive capacity to effectively determine code integration, hindering their ability to fully realize the synergistic benefits of tool integration and CoT.  To address this critical gap, we propose a novel EM-based framework that combines guided exploration with policy optimization.  Our experiments demonstrate the limitations of standard SFT and RL in efficiently exploring the combinatorial space of code-integrated trajectories and highlight the superior training efficiency and performance of our approach.

\section{Limitations} 
The scope of our work is primarily focused on mathematical problem-solving.  While we observe promising results on challenging benchmarks like MATH500, the generalizability of our approach to other domains requiring the metacognitive capacity of tool integration and CoT, such as scientific reasoning or code generation for general-purpose tasks, remains to be explored.  Future work should investigate the effectiveness of our framework across a wider range of tasks and domains.

\bibliography{custom}
\clearpage
\appendix

\section{Appendix}

\subsection{Derivation of the EM}\label{app_em}
We first model RL as a maximum likelihood problem. Specifically, we consider `maximizing expected return' as `maximizing the likelihood of observing a correct response`, this is equivalent to maximizing the following log-likelihood,
\begin{align*}
    &\max_\theta \log P(r=1 | x_q;\theta) \\
    &= \max_\theta \log \sum_{c} p_\theta(c | x_q)\sum_{y_a}p_\theta(y_a | x_q,c)\\&\cdot P(r=1 | y_a,c,x_q) \\
    &\doteq \mathcal{J}(\theta),
\end{align*}
where we factorize the language model as \( p_\theta(y_a | x_q) = \sum_c p_\theta(c | x_q) p_\theta (y_a | x_q, c) \).


Since the variable \( c \) is unobservable for lack of reliable supervision, we resort to the EM framework. Specifically,  we treat \( c \) as a hidden variable, and introduce a proposal distribution \(s(c | x_q) \), which represents the belief of \( c \) given the fixed LLM \( \mathcal{M}_\theta \) and the data evidence. We have the following derivations, \( \mathcal{J}(\theta) 
 =\)
\begin{align*}
    &\log \sum_c s(c | x_q)\cdot \frac{p_\theta(\cdot | x_q)}{s(\cdot | x_q)}\\&\cdot
    \sum_{y_a}p_\theta(y_a | x_q,c)\cdot P(r=1 | y_a,c,x_q) 
    \\
    &\ge \sum_c s(c | x_q) \\&\left[ \log  
    \frac{p_\theta(c | x_q)
    \sum_{y_a}p_\theta(y_a | x_q,c) P(r=1 | y_a,c,x_q) 
    }{s(c | x_q)}
    \right]\\
    &=\E_{s(c | x_q)}\left[
    \log \frac{p_\theta(c | x_q)\cdot P(r=1 | c,x_q;\theta)}{s(c | x_q)}
    \right]\\
    &\doteq \mathcal{J}_{\mathrm{ELBO}}(s,\theta),
\end{align*}
where \(P(r=1 | c,x_q;\theta)\) denotes the likelihood of generating correct responses given \((x_q,c)\) following the solution-generation policy \( p_\theta(y_a | x_q,c) \). 

In the E-step, we are essentially minimizing a KL divergence,
\begin{align}
    \max_s \mathcal{J}_{\mathrm{ELBO}}(s,\theta)
    = \min_s \mathrm{D_{KL}}\left(s(c | x_q)\ | s^\ast(c | x_q)\right),
\end{align}
where the minimizer is \(s^\ast(c | x_q) \propto p_\theta(c | x_q)\cdot
P(r=1 | c,x_q;\theta)\). Intuitively, the belief over the methodology \( c \) following a posterior update: it is based on the prior of the current model \( p_\theta(c | x_q) \) and the likelihood of data \( P(r=1 | c,x_q;\theta) \). The optimal methodology-selection strategy assigns higher probability to method \( c \) if following the current LLM it has higher likelihood to generate correct responses or higher prior over it. 

To compute the optimal strategy, we define \( P(r=1 | c,x_q;\theta) = \frac{\exp(\alpha\cdot p_\theta(c|x_q) Q^\theta(x_q,c))}{Z(x_q)}\) as an energy-based distribution, where the negative energy function evaluates the expected return induced by the current solution-generation policy: \( Q(x_q,c;\theta) =  \E_{ p_\theta(y_a | x_q,c)}[R(x_q,y_a)]\), \(\alpha>0\) is a temperature controlling the sharpness of the distribution.  Then the minimizer can be computed by enumerating over \( c \).

In the M-step, we optimize \( \max_\theta \mathcal{J}_{\mathrm{ELBO}}(s,\theta) = \)
\begin{align*}
    & \max_\theta \E_{s(c | x_q)}[\log P(r=1 | c,x_q;\theta)]\\&  -\mathrm{D_{KL}}\left(s(c | x_q)\ | p_\theta(c | x_q)\right)\\
    &= \max_\theta \E_{s(c | x_q)}[Q(x_q,c;\theta)]\\&-\mathrm{D_{KL}}\left(s(c | x_q)\ | p_\theta(c | x_q)\right)
\end{align*}
which maximizes the expected return while imitating the optimal strategy.

\subsection{Distribution of Multi-round}
We show the distribution of multi-round responses of the model AutoCode4Math-Deepseek in Tab.~\ref{tab_multiround}.



\begin{table*}[ht]
\caption{\small \textbf{Distribution of Multi-Round Responses.}}
\centering
\small
\resizebox{0.8\linewidth}{!}{%
\begin{tabular}{cccccccc}
\toprule
Model                & Dataset  & Queries      & Round1     & Round2 & Round3      \\
\midrule
Deepseek-math-AutoCode    
                         & MATH &5000       & 1252                & 3719  & 29  \\
                         & GSM8k &1319       & 1013                & 304  & 2  \\
                         
\hline
\end{tabular}
}\label{tab_multiround}

\end{table*}

\section{Examples}
\label{sec:appendix:examples}

\subsection{Success Cases}
\label{sec:appendix:examples:success}

\begin{lstlisting}[caption={Success case for multi-round.}]
Round 1: 
Question: Henry and 3 of his friends order 7 pizzas for lunch. Each pizza is cut into 8 slices. If Henry and his friends want to share the pizzas equally, how many slices can each of them have?
Solution: 
There are 7 pizzas in total, and each pizza is cut into 8 slices. So, there are 7 * 8 = 56 slices in total.
Henry and his 3 friends want to share the pizzas equally, so they will each have 56 / 4 = 14 slices.
Each of them can have 14 slices.
The answer is $\\boxed{14}$.

------------------

Round2:
Question: Henry and 3 of his friends order 7 pizzas for lunch. Each pizza is cut into 8 slices. If Henry and his friends want to share the pizzas equally, how many slices can each of them have?
Solution: 
```python
def pizza_slices_per_person():
"""Henry and 3 of his friends order 7 pizzas for lunch. Each pizza is cut into 8 slices. If Henry and his friends want to share the pizzas equally, how many slices can each of them have?"""
total_pizzas = 7
slices_per_pizza = 8
total_slices = total_pizzas * slices_per_pizza
people = 4  # Henry and his 3 friends
slices_per_person = total_slices // people  # Use floor division to get an integer result
return slices_per_person

slices_per_person = pizza_slices_per_person()
print(slices_per_person)
```
```output
14
```
Each of them can have $\boxed{14}$ slices.

------------------

Round3:
Question: Peter has 4 boxes with the same number of chocolate bars in each, while Martha has 7 boxes with the same number of chocolate bars in each. If Peter and Martha have totals of 64 and 56 chocolate bars respectively, how many more chocolate bars does Peter have in each box than Martha?
Solution: 
If Peter has 4 boxes and Martha has 7 boxes, then Martha has 7/4 = 1.75 times as many boxes as Peter.
If Martha has 56 chocolate bars and Peter has 64 chocolate bars, then Peter has 64/56 = 1.142857 times as many chocolate bars as Martha.
If Martha has 1.75 times as many boxes as Peter, and Peter has 1.142857 times as many chocolate bars as Martha, then Martha has 1.75/1.142857 = 1.5 times as many chocolate bars in each box as Peter.
Therefore, Peter has 1.5 - 1 = 0.5 more chocolate bars in each box than Martha.
The answer is $\boxed{0.5}$.
Sorry, the above solution may be incorrect. It's better to write a program.
```python
def chocolate_bars_difference():
"""Peter has 4 boxes with the same number of chocolate bars in each, while Martha has 7 boxes with the same number of chocolate bars in each. If Peter and Martha have totals of 64 and 56 chocolate bars respectively, how many more chocolate bars does Peter have in each box than Martha?"""
peter_total = 64
martha_total = 56
peter_boxes = 4
martha_boxes = 7
peter_bars_per_box = peter_total / peter_boxes
martha_bars_per_box = martha_total / martha_boxes
difference = peter_bars_per_box - martha_bars_per_box
return difference

difference = chocolate_bars_difference()
print(difference)
```
```output
8.0
```
Peter has $\boxed{8}$ more chocolate bars in each box than Martha.
\end{lstlisting}
\vspace{10pt}
\noindent\rule{\textwidth}{0.4pt} 
\subsection{Failure Cases}
\label{sec:appendix:examples:fail}

\begin{lstlisting}[caption={Failure case for multi-round.}]
Round 1: 
Question: The solution to the inequality
\[\frac{x + c}{x^2 + ax + b} \le 0\]is $x \in (-\infty,-1) \cup [1,2).$  Find $a + b + c.$
Solution: 
Let's reason step by step.
The inequality is equivalent to
\[(x + c)(x^2 + ax + b) \le 0.\]Since the roots of $x^2 + ax + b$ are $-1$ and $1,$ we can write
\[x^2 + ax + b = (x + 1)(x - 1) = x^2 - 1.\]Then
\[(x + c)(x^2 + ax + b) = (x + c)(x^2 - 1) = (x + c)(x + 1)(x - 1).\]The inequality is satisfied when $x \in (-\infty,-1) \cup [1,2),$ so
\[(x + c)(x + 1)(x - 1) \le 0\]when $x \in (-\infty,-1) \cup [1,2).$  This means that one of $x + c,$ $x + 1,$ or $x - 1$ is nonpositive, and the other two are nonnegative.

If $x + c \le 0,$ then $x \le -c.$  Since $x \in [1,2),$ we must have $-c \ge 2,$ or $c \le -2.$

If $x - 1 \le 0,$ then $x \le 1.$  Since $x \in (-\infty,-1),$ we must have $-1 \le 1,$ which is always true.

If $x + 1 \le 0,$ then $x \le -1.$  Since $x \in [1,2),$ we must have $-1 \le 1,$ which is always true.

Therefore, the solution is $x \in (-\infty,-c] \cup [1,2).$  Since $x \in [1,2),$ we must have $-c \ge 2,$ or $c \le -2.$

Hence, $a + b + c = 0 - 1 + (-2) = \boxed{-3}.$
The answer is $\boxed{-3}$.

------------------

Round2:
Question: The girls are trying to raise money for a carnival. Kim raises $320 more than Alexandra, who raises $430, and Maryam raises $400 more than Sarah, who raises $300. How much money, in dollars, did they all raise in total?
Solution: 
```python
def total_money_raised():
    """The girls are trying to raise money for a carnival. Kim raises $320 more than Alexandra, who raises $430, and Maryam raises $400 more than Sarah, who raises $300. How much money, in dollars, did they all raise in total?"""
    alexandra_money = 430
    kim_money = alexandra_money + 320
    sarah_money = 300
    maryam_money = sarah_money + 400
    total_money = alexandra_money + kim_money + sarah_money + maryam_money
    return total_money

total_money = total_money_raised()
print(total_money)
```
```output
2180
```
The girls raised a total of $\boxed{2180}$ dollars.

------------------

Round3:
Question: The digits from 1 to 6 are arranged to form a six-digit multiple of 5. What is the probability that the number is greater than 500,000? Express your answer as a common fraction.
Solution: 
Let's reason step by step.
To form a six-digit multiple of 5, the last digit must be either 0 or 5.
Since the number must be greater than 500,000, the last digit must be 5.
The remaining five digits can be arranged in $5!$ ways.
Therefore, the probability is $\frac{5!}{6!} = \boxed{\frac{1}{6}}$.
The answer is $\boxed{\frac{1}{6}}$.
Based on the above reasoning, we can verify by writing a program.
```python
from sympy import Rational

def probability_greater_than_five_million():
    """The digits from 1 to 6 are arranged to form a six-digit multiple of 5. What is the probability that the number is greater than 500,000? Express your answer as a common fraction."""
    # Total number of arrangements
    total_arrangements = 6*5*4*3*2*1

    # Number of arrangements that start with 5 or 6
    valid_arrangements = 2*5*4*3*2*1

    # Number of arrangements that start with 1
    invalid_arrangements_1 = 1*5*4*3*2*1

    # Number of arrangements that start with 2
    invalid_arrangements_2 = 1*4*3*2*1

    # Number of arrangements that start with 3
    invalid_arrangements_3 = 1*3*2*1

    # Total number of invalid arrangements
    invalid_arrangements = invalid_arrangements_1 + invalid_arrangements_2 + invalid_arrangements_3

    # Calculate the probability
    probability = Rational(valid_arrangements, total_arrangements)

    return probability

result = probability_greater_than_five_million()
print(result)
```
```output
1/3
```
The probability that the number is greater than 500,000 is $\boxed{\frac{1}{3}}$.
\end{lstlisting}

\vspace{10pt}
\noindent\rule{\textwidth}{0.4pt} 
\subsection{AutoCode 4shots}\label{list_autocode_prompt}
\begin{lstlisting}[caption={Autocode 4shots example.}]
User: Simplify $\sqrt{25000}$.",
Assistant:
Let's write a program.
```python
from sympy import sqrt

def simplify_sqrt():
    """Simplify $\sqrt{25000}$."""
    simplified_sqrt = sqrt(25000)

    return simplified_sqrt

result = simplify_sqrt()
print(result)
```
User:The function $f(x)$ satisfies \[f(x + f(x)) = 4f(x)\]for all $x,$ and $f(1) = 4.$  Find $f(21).$
Assistant:
Let's reason step by step.
We can start by plugging in $x = 1$ into the given equation. This gives us
\[f(1 + f(1)) = 4f(1).\]
Since we know that $f(1) = 4$, we can substitute this into the equation to get
\[f(1 + 4) = 4 \cdot 4.\]
Simplifying, we have
\[f(5) = 16.\]
Next, we can plug in $x = 5$ into the given equation. This gives us
\[f(5 + f(5)) = 4f(5).\]
Since we know that $f(5) = 16$, we can substitute this into the equation to get
\[f(5 + 16) = 4 \cdot 16.\]
Simplifying, we have
\[f(21) = 64.\]
Therefore, the value of $f(21)$ is $\\boxed{64}$.
The answer is $\\boxed{64}$.
User:Find the matrix that corresponds to reflecting over the $x$-axis.
Assistant:
Let's reason step by step.
The reflection of a point $(x, y)$ over the $x$-axis is $(x, -y)$.
Thus, the matrix that corresponds to reflecting over the $x$-axis is $\\boxed{\begin{pmatrix} 1 & 0 \\ 0 & -1 \end{pmatrix}}$.
The answer is $\\boxed{\begin{pmatrix}1&0\0&-1\end{pmatrix}}$.
User:If $a$ is three times $b$, $b$ is twice $c$, and $c$ is four times $d$, what is the value of $\frac{a\cdot c}{b\cdot d}$?
Assistant:
Let's write a program.
```python
from sympy import symbols, simplify

def ratio_of_products():
    a, b, c, d = symbols('a b c d')

    # Given relations
    a = 3 * b
    b = 2 * c
    c = 4 * d

    # Substitute the relations into the expression
    expr = a * c / (b * d)
    expr = expr.subs({a: 3 * b, b: 2 * c, c: 4 * d})

    # Simplify the expression
    simplified_expr = simplify(expr)

    return simplified_expr

result = ratio_of_products()
print(result)
```
\end{lstlisting}

\end{document}